\documentclass[a4paper]{llncs}

\usepackage{amsmath}
\usepackage{mathtools}
\usepackage{graphicx}
\usepackage[skip=0.333\baselineskip]{caption}
\usepackage{subcaption}
\usepackage{caption}
\usepackage{paralist}
\usepackage{xspace}
\usepackage{svg}
\usepackage{cite}

\usepackage{tikz}
\usetikzlibrary{shapes.geometric, arrows}

\tikzstyle{startstop} = [rectangle, rounded corners, 
minimum width=3cm, 
minimum height=1cm,
text centered, 
draw=black, 
fill=red!30]

\tikzstyle{io} = [trapezium, 
trapezium stretches=true, 
trapezium left angle=70, 
trapezium right angle=110, 
minumum width=2cm, 
minimum height=1cm, text centered, 
draw=black, fill=blue!30]

\tikzstyle{process} = [rectangle, 
minimum width=3cm, 
minimum height=1cm, 
text centered, 
text width=3cm, 
draw=black, 
fill=orange!30]

\tikzstyle{decision} = [diamond, 
minimum width=3cm, 
minimum height=1cm, 
text centered, 
draw=black, 
fill=green!30]
\tikzstyle{arrow} = [thick,->,>=stealth]

\usepackage{listings}
\usepackage{xcolor}

\definecolor{codegreen}{rgb}{0,0.6,0}
\definecolor{codegray}{rgb}{0.5,0.5,0.5}
\definecolor{codepurple}{rgb}{0.58,0,0.82}
\definecolor{backcolour}{rgb}{0.95,0.95,0.92}

\lstdefinestyle{mystyle}{
    backgroundcolor=\color{backcolour},   
    commentstyle=\color{codegreen},
    keywordstyle=\color{magenta},
    numberstyle=\tiny\color{codegray},
    stringstyle=\color{codepurple},
    basicstyle=\ttfamily\footnotesize,
    breakatwhitespace=false,         
    breaklines=true,                 
    captionpos=b,                    
    keepspaces=true,                 
    numbers=left,                    
    numbersep=5pt,                  
    showspaces=false,                
    showstringspaces=false,
    showtabs=false,                  
    tabsize=2
}

\begin{document}
    
\title{On Reducing Undesirable Behavior in Deep Reinforcement Learning Models}
\author{Ophir M. Carmel and Guy Katz}

\institute{
  The Hebrew University of Jerusalem, Jerusalem, Israel \\
  \email{ \{ophir.carmel, g.katz\}@mail.huji.ac.il} }

\maketitle

\begin{abstract}
  Deep reinforcement learning (DRL) has proven extremely useful in a
  large variety of application domains. However, even successful
  DRL-based software can exhibit highly undesirable behavior. This is
  due to DRL training being based on maximizing a reward function,
  which typically captures general trends but cannot precisely
  capture, or rule out, certain behaviors of the system. In this
  paper, we propose a novel framework aimed at drastically reducing
  the undesirable behavior of DRL-based software, while maintaining
  its excellent performance. In addition, our framework can assist
  in providing engineers with a comprehensible characterization of
  such undesirable behavior. Under the hood, our approach is based on
  extracting decision tree classifiers from erroneous state-action
  pairs, and then integrating these trees into the DRL training loop,
  penalizing the system whenever it performs an error.  We provide a
  proof-of-concept implementation of our approach, and use it to
  evaluate the technique on three significant case studies.  We find
  that our approach can extend existing frameworks in a
  straightforward manner, and incurs only a slight overhead in
  training time. Further, it incurs only a very slight hit to
  performance, or even in some cases --- improves it, while
  significantly reducing the frequency of undesirable behavior.
\end{abstract}

\section{Introduction}

Deep reinforcement learning (DRL) is a paradigm for training deep
neural network models through the application of reinforcement
learning.  DRL has proven remarkably powerful in settings involving
sequential decision making, such as game
playing~\cite{MnKaSiRuVeBeGrRiFiOs15,YeChZhChYuLiCh20,LaCh17},
Internet congestion control algorithms~\cite{JaRoGoScTa19}, and smart
transportation systems~\cite{KaBaDiJuKo17,Pa20}. This trend is likely
to intensify in coming years, with DRL taking part in an increasing
number of mission-critical software systems.

Despite its impressive success, DRL has a significant drawback.  As
with other deep-learning-based methods, DRL models are \emph{opaque}:
it is highly difficult for humans to comprehend their internal
decision making, and consequently to guarantee that the resulting
software is error-free. This is not merely a theoretical issue:
several undesirable behaviors have been observed in modern DRL-based
software (e.g.,~\cite{ElKaKaSc21} and~\cite{KaBaKaSc19}). If these
issues are not addressed, they could hamper the deployment of DRL
models in various domains of interest.

It is generally accepted that many bugs, or inaccuracies, in DRL
models stem from the \emph{reward function} in
use~\cite{AmOlStChScMa16}.  In DRL, the reward function is the
objective that the model is trained to optimize; and so, the resulting
model can only be as good as its reward function. In complex systems,
there are usually multiple goals to be satisfied simultaneously,
leading to complex reward functions. Further, there are often multiple
policies that can achieve high rewards, meaning that two models that
display similar performance (i.e., achieve similar reward scores) may
be quite different from each other, making them difficult to
compare. Finally, even ``good'' models may present highly undesirable
behavior --- in cases that were not adequately addressed in the reward
function. For example, it has been observed that the Aurora congestion
control system~\cite{JaRoGoScTa19} might sometimes choose to increase
a sender's sending rate over an already congested network; or choose
to decrease it even when the current bandwidth is extremely
under-utilized~\cite{ElKaKaSc21}. Both of these actions are clearly
incorrect, even though the Aurora model is generally very successful.

Here, we present a novel engineering methodology for reducing, and
sometimes nearly preventing, such undesirable behavior in DRL-based
software. Our approach uses a \emph{reward reshaping}
technique~\cite{Wi10}: it influences the DRL reward function in subtle
ways, in order to eventually produce a model that satisfies the
original requirements, but which is also less likely to produce
undesirable behavior. The key novelty in our approach is the way bad
behavior is expressed. Starting with an initial, already-trained model
we
\begin{inparaenum}[(i)]
\item collect instances where the system presented undesirable
  behavior;
\item generate from these state-action pairs a \emph{decision tree}
  that expresses an infinite set of undesirable behaviors; and finally
\item \emph{inject} this decision tree back into the training process,
  generating an augmented model that is far more likely to operate
  correctly compared to the original.
\end{inparaenum}

Our approach includes several novel aspects, that differentiate it
from existing techniques. Most notably, most existing techniques for
\emph{safe reinforcement learning} (e.g.~\cite{Da18,TeMaMa18}) rely
on the assumption that the undesirable behaviors are known, and are
straightforward to manually specify. We argue that this is often not
the case --- and that in fact it may be quite difficult for engineers
and stakeholders to pinpoint the root case of the undesirable
behavior. Our method addresses this difficulty by placing only a
minimal burden on the humans in the loop, namely the labeling of
state-action pairs as good or bad. In fact, our method has the added
benefit that the resulting decision trees can be used to
\emph{explain} to the engineers what went wrong. Further, our approach
does not require the resulting DRL model to globally satisfy some hard
constraints, which may be infeasible~\cite{CaMaSeLeVa23}; instead, we
offer users a trade-off between accuracy and safety, and allow them to
fine-tune it according to their specific needs.  Also, our approach is
fairly simple to understand and implement, and can extend a variety of
existing frameworks, as we later demonstrate,

For evaluation purposes, we apply our approach to three diverse case
studies: \emph{Aurora}, \emph{Traffic Control} and \emph{Snake}. In
two of these case studies, our approach succeeds in decreasing the
frequency of undesirable behavior significantly, while maintaining
(and even improving) performance, and without substantially increasing
the required training time. In the third case study, we also succeed
in decreasing the frequency of undesirable behavior significantly,
albeit while slightly degrading performance. Our success in tackling
these three, very different systems, showcases the wide applicability
of our approach.

The rest of this paper is organized as follows. In
Sec.~\ref{sec:background} we present the necessary background on DRL
and decision trees, and in Sec.~\ref{sec:approach} we describe the
different steps of our approach. In Sec.~\ref{sec:case studies} we
describe our three case studies, and then describe how our approach
was applied to them in Sec.~\ref{sec: result analysis}. Related work
is discussed in Sec.~\ref{sec:related}, and we conclude with
Sec.~\ref{sec:conc}.

\section{Background}
\label{sec:background}

\subsection{Supervised Learning and Decision Trees}
\begin{figure}[h]
\centering
\includesvg[width=0.9\textwidth]{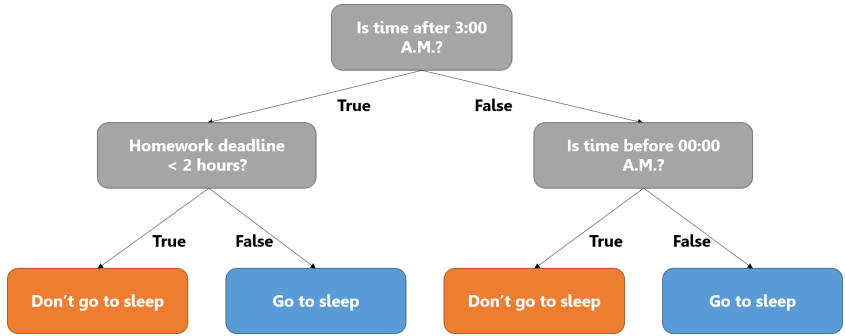}
\caption{ An toy example of a decision tree classifier, with 2 labels:
  ``Go to Sleep'' and ``Don't go to sleep''.}
\label{fig:decision-tree-example}
\end{figure}

Supervised learning \cite{LiWu12} is a machine-learning paradigm, in
which labeled samples are generalized into a function that maps
previously unseen inputs into a set of outputs. In classification
tasks, the set of outputs is finite.  Decision trees
\cite{SaLa91,KiSa08,KiSa08} are a particular kind of classifiers,
which resemble binary trees. Each tree node represents a query about
the value of some data feature; and it splits the data into two sets,
depending on whether the query evaluates to true or false (see
Fig.~\ref{fig:decision-tree-example}). In a ``good'' decision tree,
each node will split the data into two sets that are of similar
cardinality, so that the tree does not become too deep. The splitting
process continues with each internal node of the tree, until reaching
a leaf, which corresponds to one of the possible output labels.

Of the many existing forms of classifiers (e.g., deep neural networks), decision trees are considered to be fairly interpretable, because they consist of a sequence of queries pertaining directly to input features~\cite{SaLa91,KiSa08}.

\subsection{Deep Reinforcement Learning}
Deep learning~\cite{LeBeHi15} is a machine learning approach for training \emph{deep neural networks} (\emph{DNNs}). In a DNN, a complex input is processed in an iterative fashion, with each layer of the DNN computing a set of latent features, using both linear and non-linear transformations. Deep learning has had amazing success in recent years, due to its uncanny ability to generalize and learn complex structures.

\begin{figure}[h]
\centering
\includesvg[width=0.5\textwidth]{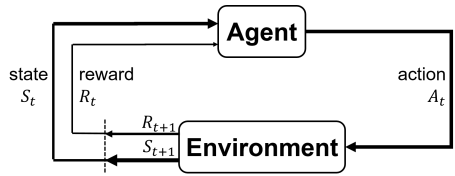}
\caption{ (Borrowed from \cite{Bh19}) The main loop of reinforcement learning. }
\label{fig:RL-loop}
\end{figure}

Reinforcement learning (RL)~\cite{SuBa18} is a machine learning
paradigm, in which an agent learns by iteratively interacting with its
environment, while trying to maximize a \emph{reward function}. In
time step $t$ of the execution, the environment is in some state
$s_t$, and the agent selects some action $a_t$. The environment then
transitions to state $s_{t+1}$, and provides the agent with a
\emph{reward} value, $r_t$, indicating how well the agent has
performed. The agent's goal is to maximize the cumulative discounted
reward; that is, to choose an action that maximizes the current
reward, and also the next rewards. The agent does this by learning a
policy $\pi$, which maps each state to the best possible action in
that state. The RL training loop is illustrated in
Fig.~\ref{fig:RL-loop}.

One issue with RL is its limited scalability: learning an optimal, or
even an approximately optimal policy has been observed to be
computationally difficult in complex systems~\cite{ArDeBrBh17}. To
overcome this limitation, engineers now apply \emph{deep reinforcement
  learning} (DRL)~\cite{Li17,ArDeBrBh17,MoScHo18,WaLiZhFeHuLiZh20},
in which the policy being learned is expressed as a DNN. This approach
has proven both scalable and powerful~\cite{WaLiZhFeHuLiZh20,ArDeBrBh17}.

\subsection{Reward Reshaping}
Reward reshaping is an RL technique, in which external knowledge from
domain experts is utilized to adjust the rewards provided to the agent
in training, in order to improve the learned
policy~\cite{HuWaJiWaChHaWuFa20,Wi10}. This technique can also
expedite training, and is particularly useful in settings where reward
accumulation is slow during early training~\cite{Wi10}. When reward
reshaping is applied, the reward $r_t$ is replaced with some modified
reward $r_t+f_t$, where $f_t$ is decided by the external expert, as
opposed to the environment. Common reward reshaping techniques seek to
improve the performance of the resulting agent~\cite{NgHaRu99} and to
reduce its training time~\cite{WiCoEl03}.

\section{Approach}\label{sec:approach}
We propose a novel reward-reshaping based approach, aimed at allowing
engineers to benefit from the advantages of DRL while reducing or
avoiding the undesirable behaviors it usually entails. The high-level
steps of our approach are:
\begin{enumerate}
\item Obtain traces of the initial DRL model, and label state-action pairs in those traces as 
  desirable or undesirable. Then, generalize these labeled pairs into a
  decision tree.
\item Manually inspect the decision tree to gain insight into the root
  cause of the problem; adjust the tree if needed.
\item Leverage the decision tree to re-train the DRL model, with a
  reshaped reward function aimed at reducing the frequency of undesirable behavior.
\item Analyze the results, possibly fine-tuning hyper-parameters if needed.
\end{enumerate}
We now proceed to explain each of these steps in greater detail.

\subsection{Detecting Undesirable Behavior}
The first step of our approach entails characterizing the undesirable
behavior of the system. Ideally, we would describe this behavior as a
logical formula, but this is known to be difficult in
practice~\cite{ClHeVeBl18}. To circumvent this difficulty, we use only
a simple form of specification: $\langle$state, action$\rangle$ pairs,
labeled by a human expert to indicate whether selecting this action in
this state is \emph{desirable} or \emph{undesirable}. 

As a running example, consider the Aurora congestion control
system~\cite{JaRoGoScTa19}. Aurora's goal is to maximize the
throughput of a computer network, by penalizing the agent when latency
is observed or when packets are lost (additional details appear in
Sec.~\ref{sec:case studies}). Consider a situation in which the agent
observes perfect network conditions, i.e. low latency and no packet
loss; and yet chooses to decrease the packet sending rate, which will
likely decrease the network's throughput. Clearly, this is undesirable
behavior. Here, we propose to rely on a human expert to mark such
state-actions pairs as undesirable, even if that expert is unable, or
unwilling, to write a logical formula that precisely captures these
cases.

Once we have a set of state-action pairs marked as desirable or
undesirable, our next step is to generalize them, through supervised
learning, so that we are able to classify additional, previously
unseen state-action pairs. We choose here to use decision trees, in
order to benefit from their relative
interpretability~\cite{KiSa08,SaLa91}. We next discuss how these trees
are trained in our setting.

\medskip
\noindent
\textbf{Grammars.}
Our choice of decision trees is geared towards improved
explainability; but in order to fully tap their potential, we need
their internal nodes to represent meaningful queries on the data,
which could then assist humans in interpreting them. Naturally,
different choices of features may be adequate for different problem
domains. Thus, we parameterize our approach with a \emph{grammar},
which defines the set of features that a tree may contain. We define
what a grammar is in our context, and propose here some grammars that
are adequate for common problem domains; and these may be fine-tuned
to support additional domains, as needed.



\begin{definition}[Grammars]\label{def:Grammars}
Observe a DRL agent, whose environment states $s_1,s_2,\ldots$ are comprised of a set of features of interest. In our context, 
    A grammar rule is a function that can be applied to these features (individual features, or sets thereof). 
    A \emph{grammar} is a set of grammar rules.
\end{definition}

The idea behind a grammar is to allow the user to provide functions and predicates relevant to the problem at hand; and then use these to construct formulas that will populate the nodes of the decision tree. We regard the grammar rules as templates, and the features as instances of those templates.
Clearly, different grammars may result in different decision trees. Even if these different trees, when integrated into the DRL training, result in similar reward values, they can afford highly varying degrees of interpretability: a ``good'' grammar, which uses
predicates that are relevant to the system at hand, will result in a
decision tree that is more straightforward for humans to comprehend
(e.g., is smaller), and will thus better contribute to the system's explainability.



As a toy example, consider a DRL agent whose input (the system state) is a vector of real-valued numbers. A reasonable grammar to attempt in this case might consist of rules (functions and predicates) such as $\geq$, $>$ and $=$, \emph{average} or \emph{max}. Some of these rules might prove less appropriate if the system states are Boolean vectors. 

\medskip
\noindent
\textbf{Training.}  Once we select a grammar, the next step is to use
it to generate a decision tree for predicting whether a state-action
pair represents undesirable behavior. Formulas produced by the grammar
serve as the features for this tree; and the actual training can be
carried out using any standard framework
(e.g.,~\emph{sklearn}~\cite{PeVaGrMiThGr11}).

\subsection{Inspecting Decision Trees}
After creating the decision tree, we propose to manually inspect it in
order to try and determine the root cause of the undesirable
behavior. This may allow a human expert to adjust and fine-tune the
tree, in case the original labeling was inaccurate, or if the
state-action pairs did not properly cover all relevant cases. In
addition, this may assist the human expert to better characterize the
undesirable behavior, and thus render the system more explainable.

In our use cases, we found it useful to inspect paths of the tree that
lead to the same result (desirable or undesirable), and deduce the
common behavior described by them. This was easier when the tree was
not overly deep (about depth 5), and so the number of paths was not
prohibitive. This fact is part of our motivation for selecting
adequate grammars, as these allow for more shallow trees without
sacrificing precision. We elaborate more on this topic in
Sec.~\ref{sec: result analysis}.



\subsection{Reducing Undesirable Behavior}
\begin{figure}[h]
\centering
\includesvg[width=0.4\textwidth]{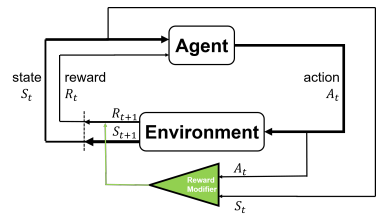}
\caption{The modified reinforcement-learning loop.}
\label{fig:new-rl-loop}
\end{figure}

Using the decision tree, we now train the model again, but with the
following modification. In each iteration of the training procedure,
after obtaining the current state, the selected action and the reward,
we evaluate the decision tree on the current state and action. If the
tree determines that the $\langle$state, action$\rangle$ pair
constitutes undesirable behavior, we modify the reward value as described next, and the
continue with the training as usual; otherwise, the reward value is 
unchanged. Fig.~\ref{fig:new-rl-loop} depicts the overall flow.

\medskip
\noindent
\textbf{Reward Modification.}  When undesirable
behavior is detected, we penalize the agent by multiplying
its reward $r$ by a \emph{reward-modifier} $\alpha$, so that $r*\alpha \le r$.  
For a positive initial reward, we set 
$\alpha \in [0,1]$; and for a negative reward, 
$\alpha\in[1,\infty]$.  Listing~\ref{code:reward_modifier} depicts the
pseudocode for the \textit{get-reward-modifier} function.


\begin{lstlisting}[language=Python, caption={The \emph{get reward modifier} function.}, label={code:reward_modifier}]
def get_reward_modifier(tree, state, action):
    outcome = tree.classify(state, action)
    return ( outcome == UNDESIRABLE_BEHAVIOR ) ? REWARD_MODIFIER : 1
\end{lstlisting}
Penalizing the reward is then performed as shown in Listing.~\ref{code:modify_reward}.

\begin{lstlisting}[language=Python, caption={The \emph{modify reward} function}, label={code:modify_reward}]
def modify_reward(reward, state, action):
    modifier = get_reward_modifier(state, action)

    if reward >= 0:
        reward = reward * modifier
    else:
        reward = reward * (1/modifier)

    return reward
\end{lstlisting}
Finally, the modified training loop appears in Listing~\ref{code:training_loop}.

\begin{lstlisting}[language=Python, caption={The modified training loop.}, label={code:training_loop}]
while step < max_steps:
    # get state
    current_state = get_state()

    # get action
    action = get_action(current_state)

    # get reward
    reward = get_reward(old_state, old_action, current_state)

    # modify the reward
    reward = modify_reward(reward, state, action)

    # update environment
    environment.step(action)

    old_state = current_state
    old_action = action
    step += 1
    
    
\end{lstlisting}

\subsection{Result Analysis and Hyper-Parameter Adjustment}
After re-training the DRL model using the decision tree, we propose to
analyze the performance of the new model, and adjust the reward
modifier value if needed. Decreasing the reward modifier generally
decreases the frequency of undesirable behavior occurring, (as we
discuss in Sec.~\ref{sec: result analysis}); however, decreasing it
too much may interfere with the training and hurt performance.
Therefore, we propose to train multiple models, using varied reward
modifiers, and then choose the one that achieves the best reward
values. This process can also be automated in a straightforward way.

\section{Case Studies}\label{sec:case studies}

For evaluation purposes, we applied our approach to three case
studies, each time following the steps described in
Sec.~\ref{sec:approach}: analyzing state-action pairs, and labeling
them as desirable or undesirable behavior; writing an appropriate
grammar for the system at hand; using the labeled state-action pairs
and grammar to train a decision tree; and then using this tree to
retrain the model. We then compared the original and re-trained
models, in order to assess the effectiveness of our method. We also
examined the decision trees to better understand and characterize the
undesirable behavior of each system.
Through our experiments, we attempted to answer the following research
questions (RQs):

\begin{enumerate}
\item \textbf{RQ1:} can our approach reduce the agent's undesirable
  behavior, without significantly harming its performance? \label{rq_2}
\item \textbf{RQ2:} can our approach be used to better explain the
  undesirable behavior of the DRL-trained agent to a
  human? \label{rq_1}
\end{enumerate}

\subsection{Case Study 1: Aurora}\label{case:Aurora}
Congestion control is the task of balancing the sending rate of
packets into a computer network, in order to minimize packet loss and
maximize throughout. A main issue in congestion control is that
bandwidth is constantly changing, and the sending rate must change
accordingly~\cite{ElKaKaSc21}.  Aurora~\cite{JaRoGoScTa19} is a
congestion controller that uses a DRL agent to govern sending
rates. In each time step, it takes three input vectors, which
information about the $t$ most recent time steps in the network (in
our case, $t=10$) --- the \emph{latency gradient}, indicating an
increase or decrease in latency; the \emph{latency ratio}, indicating
the ratio between current latency and minimum latency; and the
\emph{sending ratio}, indicating the ratio between sent packets and
received packets.  Consequently, two consecutive Aurora system states
$s_1\rightarrow s_2$ share the same input vectors, except for the
oldest entry (in each vector) in $s_1$, which is dropped and replaced
with a fresh entry in $s_2$. We refer to this relation between
consecutive states, which is quite common in computer network
systems~\cite{ElKaKaSc21}, as a \emph{sliding window}.  The output of
the model is a single value, indicating whether the sending rate
should increase (positive value) or decrease (negative value), and by
how much.


\medskip
\noindent
\textbf{Grammar.}
We specify the following, "sliding window" grammar rules, for some Aurora state $s = [s_i, s_{i+1},...,s_{i+k}]$:

\begin{itemize}

\item {\textbf{Value:}}
    returns the value $\text{Value}_{i}$, defined as entry $s_i$.
    
\item {\textbf{Diff:}}
    returns the value $\text{Diff}_{i,j}$, defined as $s_i - s_j$.

\item {\textbf{Sign:}}
    returns the value $\text{Sign}_{i,j}$, defined as $Sign(s_i - s_j)$: the sign of the difference between the $i$'th and $j$'th values.

\item{\textbf{Average:}}
    returns the value $\text{Average}_{i_1,\ldots,i_m}$, defined as $(\sum_{n\in\{i_1,\ldots,i_m\}}s_{n}) / m$, i.e. the average of the values with indices $\{i_k|k\in\{0,\ldots,m\}\}$.

\item{\textbf{Action:}}
    returns the value $\text{Action}$, which is the output selected by the agent.

\end{itemize}


The motivation is that for sliding window inputs, it is useful to
inspect how inputs change over time; and that comparing directly
adjacent temporal readings is more useful than comparing those that
are far apart. We selected the following instantiations of these
rules, instantiated on the \emph{sending ratio} part of the system
state, to serve as the features for the decision tree:
\begin{align*}
&\{\textbf{Value}_i | i\in\{0,\ldots,9\}\} \cup
\{\textbf{Diff}_{i,i+1} | i\in\{0,\ldots,8\}\} \cup \{\textbf{Diff}_{i,i+2} | i\in\{0,\ldots,7\}\} \cup\\
&\{\textbf{Sign}_{i,i+1} | i\in\{0,\ldots,8\}\} \cup \{\textbf{Sign}_{i,i+2} | i\in\{0,\ldots,7\}\} \cup\\
&\{\textbf{Average}_{i,i+1} | i\in\{0,\ldots,8\}\} \cup
\{\textbf{Average}_{i,i+2} | i\in\{0,\ldots,7\}\} \cup\\
&\{\textbf{Average}_{i,i+1,i+2} | i\in\{0,\ldots,7\}\} \cup
\{\textbf{Action}\}
\end{align*}


\medskip
\noindent
\textbf{Undesirable Behavior.}
The undesirable behavior that we target is cases in which
 network conditions are nearly perfect (low latency, and almost no packet loss), but in which the agent decides to decrease the sending rate. Such actions are clearly not optimal with respect to the goal of maximizing throughput~\cite{ElKaKaSc21}. In order to detect instances of this behavior and label it, we looked for traces with very low latency and close-to-optimal sending ratios (less than 1.2 sending ratio in all 10 entries).

\subsection{Case Study 2: Traffic Control}\label{case:TrafficControl}
In \emph{Traffic Control}~\cite{Vi19}, a DRL agent manages a
road intersection. At each time step, the agent needs to determine
what the traffic lights at the intersection should show, in order to
maximize the intersection's throughput.

\begin{figure*}[h]
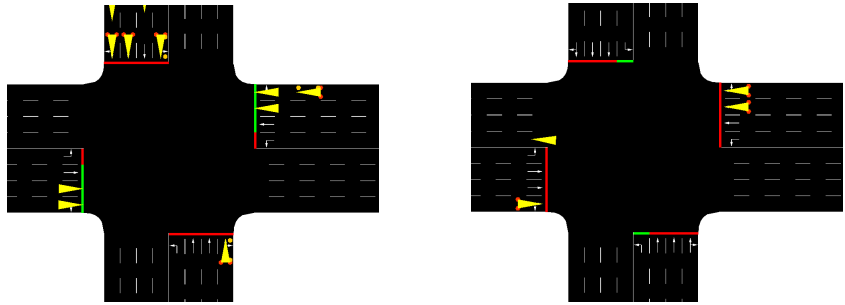

\centering
\begin{subfigure}{.5\textwidth}
  \centering
  \includesvg[width=.8\linewidth]{Images/Traffic/other/EW_phase.svg}
  \caption{An EW traffic light configuration.}
  \label{fig:traffic-phase-ew}
\end{subfigure}%
\begin{subfigure}{.5\textwidth}
  \centering
  \includesvg[width=.8\linewidth]{Images/Traffic/other/NSL_phase_new.svg}
  \caption{An NSL traffic light configuration.}
  \label{fig:traffic-phase-nsl}
\end{subfigure}
\caption{The intersection for the \emph{Traffic Control} system.}
\label{fig:traffic-zoom-in}
\end{figure*}

The intersection has eight lanes:
\begin{inparaenum}[(i)]
    \item West to East (W2E);
    \item West to Turn Left (W2TL);
    \item East to West (E2W);
    \item East to Turn Left (E2TL);
    \item North to South (N2S);
    \item North to Turn Left (N2TL);
    \item South to North (S2W); and
    \item South to Turn Left (S2TL).
\end{inparaenum}
Each lane is divided into 10 segments in different distances from the
intersection. The agent's input is 80 Boolean flags, one per segment,
indicating whether or not cars  are currently in that
segment.
The agent's output action determines which of the lanes get a green
light for this time step, and which do not, according to one of 4
possible configurations:
\begin{inparaenum}[(i)]
    \item North and South (NS), which opens up the N2S and S2N lanes;
    \item North and South Turn Left (NSL), which opens up the N2TL and S2TL lanes;
    \item East and West (EW), which opens up the E2W and W2E lanes; or
    \item East and West Turn Left (EWL), which opens up the E2TL and W2TL lanes.
\end{inparaenum}
An illustration appears in Fig.~\ref{fig:traffic-zoom-in}.

\medskip
\noindent
\textbf{Grammar.}
This use case, too, fits the sliding window category: in each lane, each car iteratively moves forward towards the intersection 
(or away from it). Since there are multiple lanes, each state consists of multiple sliding windows. Unlike
in Aurora, here each car can traverse several segments in a single time step, and so the sliding window can shift by multiple entries (as opposed to a single entry in Aurora). To accommodate this, we adjust our sliding window
grammar and compute averages and differences on the individual sliding
windows (corresponding to individual lanes). Further, to simplify the decision tree we 
remove some of the features that are less likely to be
meaningful --- namely, those that concern segments that are very far
away from the intersection, and those that concern sets of segments that are far apart from each other. The resulting grammar that we use is thus:

\begin{itemize}
\item {\textbf{Value:}}
    returns the value $\text{Value}_{lane,i}$, defined as $s_{lane, i}$ in a specific lane.

\item {\textbf{Diff:}}
    returns the value $\text{Diff}_{lane,i,j}$, defined as $s_{lane,i} - s_{lane,j}$ in a specific lane.

\item{\textbf{Average:}}
    returns the value $\text{Average}_{lane,i_1,\ldots,i_m}$, defined as $(\sum_{n\in\{i_1,\ldots,i_m\}}s_{lane,n}) / m$, i.e. the average of the values in the set of indices $\{i_k|k\in\{0,\ldots,m\}\}$ in a specific lane.

\item {\textbf{IsAction:}}
    returns the value $\text{IsAction}_{d}$, defined as $action == d$.
\end{itemize}
We use the Diff rule only for segments that are adjacent, and include the Average rule only for consecutive segments adjacent to the intersection. The instantiations of the grammar rules in this case are thus:
\begin{align*}
    &\{\textbf{Value}_{lane, i}\ |\ i\in\{0,\ldots,9\}, lane\in\ \text{LANES}\} \cup\\
    &\{\textbf{Diff}_{lane, i, i+1}\ |\ i\in\{0,\ldots,9\}, lane\in\
      \text{LANES}\} \cup\\
    &\{\textbf{Average}_{lane, 0, \ldots, m}\ |\ m\in\{1,\ldots,7\}, lane\in\ \text{LANES}\} \cup\\
    &\{\textbf{Action}_d\ |\ d\in\text{ACTIONS}\} 
\end{align*}
where
\sloppy
 $\text{LANES}=\{N2S, S2N, E2W, W2E, N2TL, S2TL,
  E2TL, W2TL\}$
 and
 $\text{ACTIONS}=\{NS,EW,NSL,EWL\}$.

\medskip
\noindent
\textbf{Undesirable Behavior.}
We target cases where some lanes are empty, but others are ``jammed'' --- that is, the two segments closest to the intersection are filled with cars --- but in which  the agent does assigns the green light to empty lanes instead of the crowded ones.  
Clearly, this choice of action is not optimal. 

\subsection{Case Study 3: Snake}\label{case:Snake}
In this case study, we study a DRL agent trained to play the game \textit{Snake}. In this
game, the agent controls a snake that is slithering around on a
board. In each time step, the snake's head can move in each cardinal direction (except the one which is the opposite of which it is currently facing); and the snake's
body always follows its head, along the exact same trajectory that the
head took. Each time, an apple will appear on the board, and the
snake's goal is to eat as many apples as possible --- without colliding
with its own tail or with a wall. Each apple collected increases the
snake's length by one, making it more challenging to maneuver
without colliding.  As part of our case study, we built on top of a
publicly available implementation of the game~\cite{Ha22}.

\begin{figure}[h]
\centering
\includegraphics[width=0.35\textwidth]{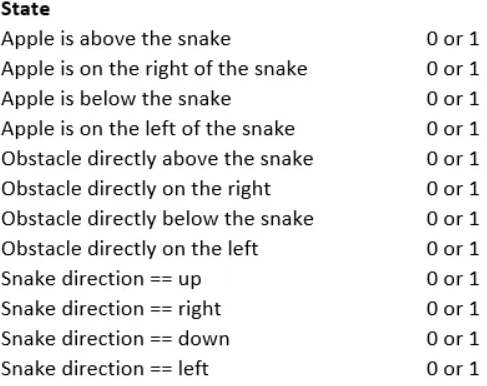}
\qquad
\includegraphics[width=0.35\textwidth]{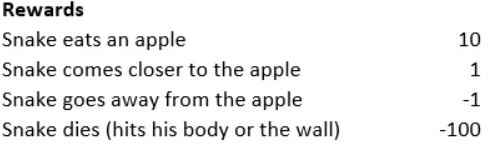}
\caption{On the left: the
  state of Snake. On the right: the rewards at each time step of
  Snake.}
\label{fig:snake_state_and_rewards_space}
\end{figure}

The input to the Snake agent is a binary array with 12 entries, with
the values depicted in
Fig.~\ref{fig:snake_state_and_rewards_space}. These entries describe
the location of the apple with respect to the snake; the location of
obstacles in the snake's vicinity; and the snake's direction. The
agent then chooses one from four possible actions: move UP, move
RIGHT, move DOWN or move LEFT.  Finally, the reward is computed as
described in Fig.~\ref{fig:snake_state_and_rewards_space}.

\medskip
\noindent
\textbf{Grammar.}
Unlike in the previous use cases, the input to the Snake agent does not include a sliding window component. Instead, it includes three sections, which are conceptually separate: 
%
%
%
%
the ``APL'' part, describing the apple's location; the ``DIR'' part, indicating the snake's direction; and the ``OBS'' part, indicating any nearby obstacles. We add another section, which is the ``ACT'' part, describing the action selected by the agent. 
Each of these parts is binary array of size 4, with each entry corresponding to a direction: up, right, down or left.
For example, state 
$
s=[1, 0, 0, 1, 0, 1, 0, 0, 1, 1, 0, 0, 0, 0, 0, 1]
$
is interpreted as 
\[
APL=[1, 0, 0, 1],\quad
DIR=[0, 1, 0, 0],\quad
OBS=[1, 1, 0, 0],\quad
ACT=[0, 0, 0, 1] 
\]
which means that the apple is in the direction ``UP-LEFT'', the snake
is facing ``DOWN'', there is an obstacle in the direction
``UP-RIGHT'', and the action selected was ``LEFT''.  We refer to this
kind of input as an \emph{equal-size partitions} input, and specify
the following grammar rules, for some Snake state
$s = [APL, DIR, OBS, ACT]$:


\begin{itemize}
\item {\textbf{Value:}}
    returns the value $\text{Value}_{part,i}$, defined as $s_{part,i}$.

\item {\textbf{IsAction:}}
    returns the value $\text{IsAction}_{d}$, defined as $s_{ACT,d} == 1$
    
\item {\textbf{IsEqual:}}
    returns the value $\text{IsEqual}_{part_1, part_2, i}$, defined as $s_{part_1, i} == s_{part_2,i}$.
\end{itemize}
where $part,part_1,part_2\in\{APL, DIR, OBS, ACT\}$.  
The instantiations of the grammar rules in this case are:
\begin{itemize}
    \item $\{
    \textbf{Value}_{part,i} \ |\
    i\in\{0,1,2,3\}, part\in\text{PARTS}\}
    \}$
    \item $\{\textbf{IsAction}_{d} \ |\ d\in\text{DIRECTIONS}\}$
    \item $\{\textbf{IsEqual}_{part1,part2,i} \ |\ i\in\{0,1,2,3\}, part1,part2\in\text{PARTS}\}$
\end{itemize}
where $\text{PARTS}=\{\text{APL, OBS, DIR, ACT}\}$ and
$\text{DIRECTIONS}=\{\text{UP, DOWN, LEFT, RIGHT}\}$.  For example,
$IsEqual_{APL, OBS, 2}$ evaluates to true when the apple and an
obstacle are to the snake's right, whereas $IsEqual_{OBS, DIR, 3}$
evaluates to true when there is an obstacle directly below the snake,
and its direction is also down.

The motivation for these rules is that in an ``equal-size partitions''
system, the inputs in the same part are less likely to have meaningful
connections, whereas inputs in different parts but which share the
same index, may present meaningful connections.  Conversely, the
sliding window rules used in Aurora and Traffic Control are mostly
irrelevant in this case. For example, the $Average_{i,j}$ rule would
compute an average over Boolean values that are most likely
independent. Instead, the proposed grammar leverages the symmetry
between the four parts of the input state, and allows inspecting the
relative directions of the snake, apple and obstacles.

\medskip
\noindent
\textbf{Undesirable Behavior.}
When examining state-action pairs, we noticed cases where the snake would already be moving towards the apple, but the agent would suddenly switch the snake to move in another direction --- even though there were no obstacles nearby. This behavior is clearly undesirable.


\section{Experiments and Results} \label{sec: result analysis}
\subsection{Experiment 1: Reduction of Undesirable Behavior}
In this experiment, we set out to answer RQ~\ref{rq_2}. To do so, we
apply our methodology to each of the case studies: we label
state-action pairs as desirable or undesirable (according to the
criteria in Sec.~\ref{sec:case studies}); train the resulting decision
trees; and then use these trees to retrain the agents. We train models
with different reward modifier values (a model trained with a reward
modifier 1.0 is just the original model), and compare their
performance. Our hypothesis is that even reward modifiers close to 1
should be sufficient to produce a significant reduction in the
frequency of undesirable behavior selected by the agent.  To measure
this, we compute the reward obtained by the modified model, and
compare it to that of the original. Further, we compute the ratio
of undesirable behaviors among all behaviors,
$\frac{\text{UNDESIRABLE}\xspace{}}{\text{DESIRABLE}\xspace{} +
  \text{UNDESIRABLE}\xspace{}}$. Additionally, we measure the time
overhead caused by our method as part of the DRL training procedure
(assuming the modified model is trained to achieve a reward similar to
that of the original).

\begin{figure}[h]
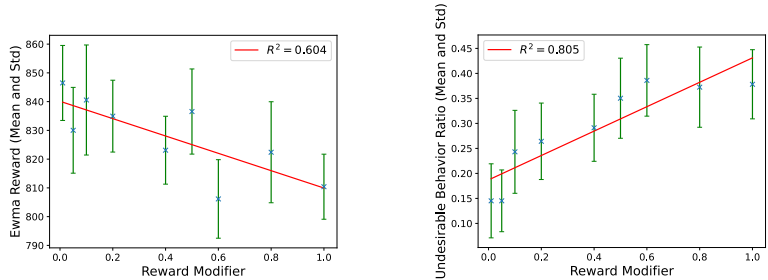

\centering
\includesvg[width=0.4\textwidth]{Images/aurora/final_results/aurora_reward_new.svg}
\qquad
\includesvg[width=0.4\textwidth]{Images/aurora/final_results/aurora_ratio_new.svg}
\caption{Aurora: on the left-hand side, the average reward values per
  reward modifiers value. On the right-hand side, the average
  undesirable behavior ratio per reward modifier value.  }
\label{fig:aurora-comparative}
\end{figure}

\medskip
\noindent
\textbf{Aurora.}  The results of applying our method to the Aurora use
case are depicted in Fig.~\ref{fig:aurora-comparative}. We trained 5
models for each of the reward modifier values
$\{1, 0.8, 0.6, 0.5, 0.4, 0.2, 0.1, 0.05, 0.01\}$. The plot on the left
shows the average running-average reward values obtained by the
models, for each reward modifier value; and the plot of the right shows
the average running-average undesirable ratio.  For training, we used
approximately 11,000 state-action pairs of desirable behavior, and an
additional 11,000 state-action pairs of undesirable behavior, recorded
during the original agent's training.  We observe that the various
agents tend to converge to a steady running-average reward level after
approximately 500 test steps. Once a model has converged, we calculated
the average reward over the next 1500 test steps, and these make up
the plot on the left.



We observe that, generally, training with small reward modifiers tends
to make agents achieve higher rewards. Also, if we examine the
training rewards, we see that the models tend to converge after a
similar number of steps. However, our approach does increase training
time by as much as 38\%, as each training iteration takes longer to
carry out.  As part of our future work we plan to reduce this
overhead, by optimizing our implementation.

The results in Fig.~\ref{fig:aurora-comparative} indicate a direct
correlation between lower reward modifier values and the scarcity of
undesirable behavior. This is expected, because lower modifier values
imply a harsher penalty to the agent for undesirable behavior. We were
able to decrease the undesirable behavior ratio by around 60\%, which
is significant; although we were not able to completely remove the
undesirable behavior, presumably because it was quite common for the
original model.

Overall, we conclude that our framework was able to achieve its
objectives for the Aurora case study: the undesirable behavior was
significantly reduced, without degrading performance --- and even
improving it in some cases.

\begin{figure}[h]
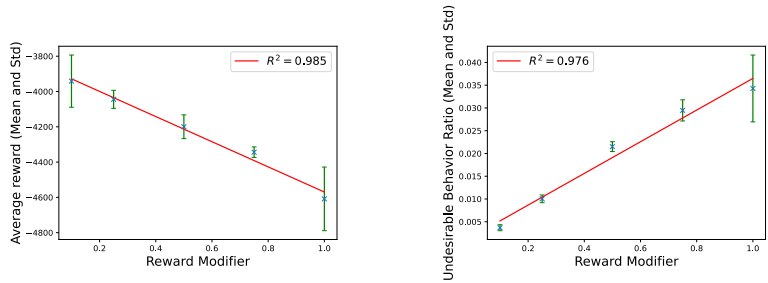

\centering
\includesvg[width=0.4\textwidth]{Images/Traffic/final_results/traffic_reward_new.svg}
\qquad
\includesvg[width=0.4\textwidth]{Images/Traffic/final_results/traffic_ratio_new.svg}
\caption{Traffic Control: on the
  left-hand side, the average reward value per reward modifier
  value. On the right-hand side, the average undesirable behavior
  ratio per reward modifier value.  }
\label{fig:traffic-comparative}
\end{figure}

\medskip
\noindent
\textbf{Traffic Control.}  Next, we performed a similar experiment for
the Traffic Control use case. We used 5 different reward modifier
values, $\{1, 0.75, 0.5, 0.25, 0.1\}$, and trained 3 different models
for each of them (with 100 training and testing iterations per model).
The left-hand side of Fig.~\ref{fig:traffic-comparative}
depicts the running-average rewards we obtained.  For training, we
used approximately 14,000 state-action pairs labeled as undesirable,
and approximately 82,000 state-action pairs labeled as desirable.
This time, the training time overhead was around 11\% for a single
episode. We also observed a difference in the number of iterations
required for convergence: it took approximately 30-40 additional
iterations for models with low reward-modifier values to converge
compared to those with higher reward-modifier values.  We observe that
in this case, applying our approach also resulted in increasing the
agent's overall reward. This is unsurprising, as the behavior we
labeled as undesirable was indeed counter productive to the agent's
goals.



Examining the ratio of undesirable behavior,
$\frac{UNDESIRABLE}{DESIRABLE + UNDESIRABLE}$, as a function of the
reward modifier (right-hand side of Fig.~\ref{fig:traffic-comparative}), we again observe a direct correlation between lower reward modifier values and the scarcity of undesirable behavior. We were able to decrease the undesirable behavior ratio by around 89.4\%, which is highly significant, and which implies that the undesirable behavior barely occurs anymore.

Overall, we conclude that our framework was able to achieve its objectives for the Traffic Control case study: the undesirable behavior was significantly reduced, without degrading performance --- and even improving it.

\medskip
\noindent
\textbf{Snake.}  Finally, we applied our method to the Snake case
study, using 5 different reward modifier values ---
$\{1, 0.75, 0.5, 0.25, 0.1\}$. We trained 4 different models for each
reward modifier value, and then performed 100 games of Snake for each
of these models.  For training, we used approximately 3,500
undesirable state-action pairs, and 18,000 desirable
pairs.  
These state-action pairs were recorded while training regular
models. The overhead in training was negligible (\~1\%).

\begin{figure}[h]
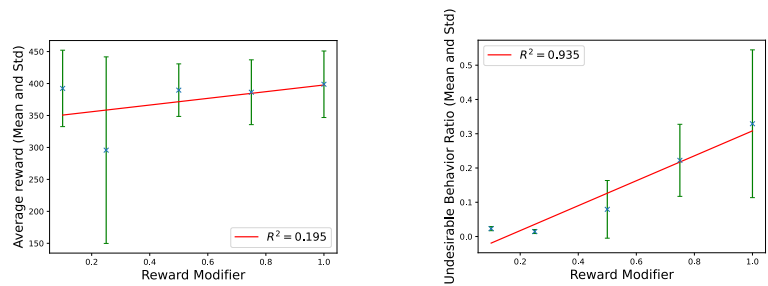

\centering
\includesvg[width=0.4\textwidth]{Images/snake/final_results/snake_reward_new.svg}
\qquad
\includesvg[width=0.4\textwidth]{Images/snake/final_results/snake_ratio_new.svg}
\caption{Snake: on the left-hand side,
  the average reward value per reward modifier value. On the
  right-hand side, the average undesirable behavior ratio per reward
  modifier value.  }
\label{fig:snake-comparative}
\end{figure}

The left-hand side of Fig.~\ref{fig:snake-comparative} demonstrates
that in this case, applying our approach actually decreased the
average reward and the performance of the model by approximately
1.5\%. This implies that if we penalize the agent too harshly (via a
very small reward modifier value), performance is actually slightly
hurt. This outlines the trade-off between performance and undesirable
behavior reduction.

Examining the ratio of undesirable behavior as a function of the
reward modifier used (right-hand side of
Fig.~\ref{fig:snake-comparative}), we once again observe a direct
correlation between lower reward modifier values and the scarcity of
undesirable behavior. We were able to decrease the undesirable
behavior ratio by around 93\%, which is highly significant, and
implies that undesirable behavior barely occurs anymore.

Overall, we conclude that our framework was able to achieve its main
objective in the Snake Control case study.

\subsection{Experiment 2: Explainability}
Another advantage of our approach is that it
can assist in \emph{explaining} the undesirable
behavior that it attempts to minimize. This is performed by presenting
the decision tree, which is already produced as part of our approach,
for manual inspection by the engineers.
In this experiment, we set out to answer RQ~\ref{rq_1}, and determine whether our approach is indeed useful in explaining
  undesirable behavior. As we later see, the answer is affirmative.
We point out that there is a certain trade-off between our two RQs:
a deeper decision tree
will likely result in higher model accuracy, but will be more
difficult for humans to manually parse; and vice versa. 
Below, we analyze this trade-off using our case studies.

\medskip
\noindent
\textbf{Aurora.}  In Aurora, the behavior we labeled as undesirable is when the
network conditions are generally good (low latency, low packet loss), but the agent chooses to decrease the sending
rate. We trained a tree over state-action pairs of desirable and undesirable behavior, which reached over 99.5\% accuracy with a depth of 9.
When inspecting this tree, we observed 2 significant paths:
\begin{itemize}
\item $ACT > 0 \rightarrow \textbf{Desirable}$. This is a significant
  path, because it implies that the selected action pays a key role in
  classification --- if the action is greater than 0, the behavior is
  always desirable. This is consistent with our targeting of cases
  where the agent would reduce the sending rate, despite network
  conditions being good.

\item
  \sloppy
  $((Value_9 < 1.195) \wedge (Value_8 < 1.195) \wedge (Value_6 <
  1.195) \wedge (Value_2 < 1.195) \wedge (AVG_{1,2,3} < 1.2))
  \leftrightarrow \textbf{Undesirable}$. In this path, if any of the
  listed clauses are False, then the behavior is desirable; and if all
  of them are True, then the behavior is undesirable. This implies
  that if the values in places 9, 8, 6 or 2 (representing past sending
  rates) are smaller than 1.195, and if the average of indices 1,2,3
  is smaller than 1.2, then the behavior is undesirable. We deduce
  from here that the tree learned that values 1.2 and 1.195 are
  significant thresholds, which separate good network conditions
  (where the sending rate should not be decreased) from bad ones. This
  is again consistent with our original labeling of traces.
\end{itemize}

Combining the information these two paths, we obtain a general and rigorous specification of the undesirable behavior --- if the agent chooses to decrease the sending rate, and also  the specified values are below the  1.2 and 1.195 threshold, then the behavior is undesirable. These values could then be inspected and fine-tuned by a domain expert. 

\begin{figure}[h]
\centering
\includesvg[width=0.9\textwidth]{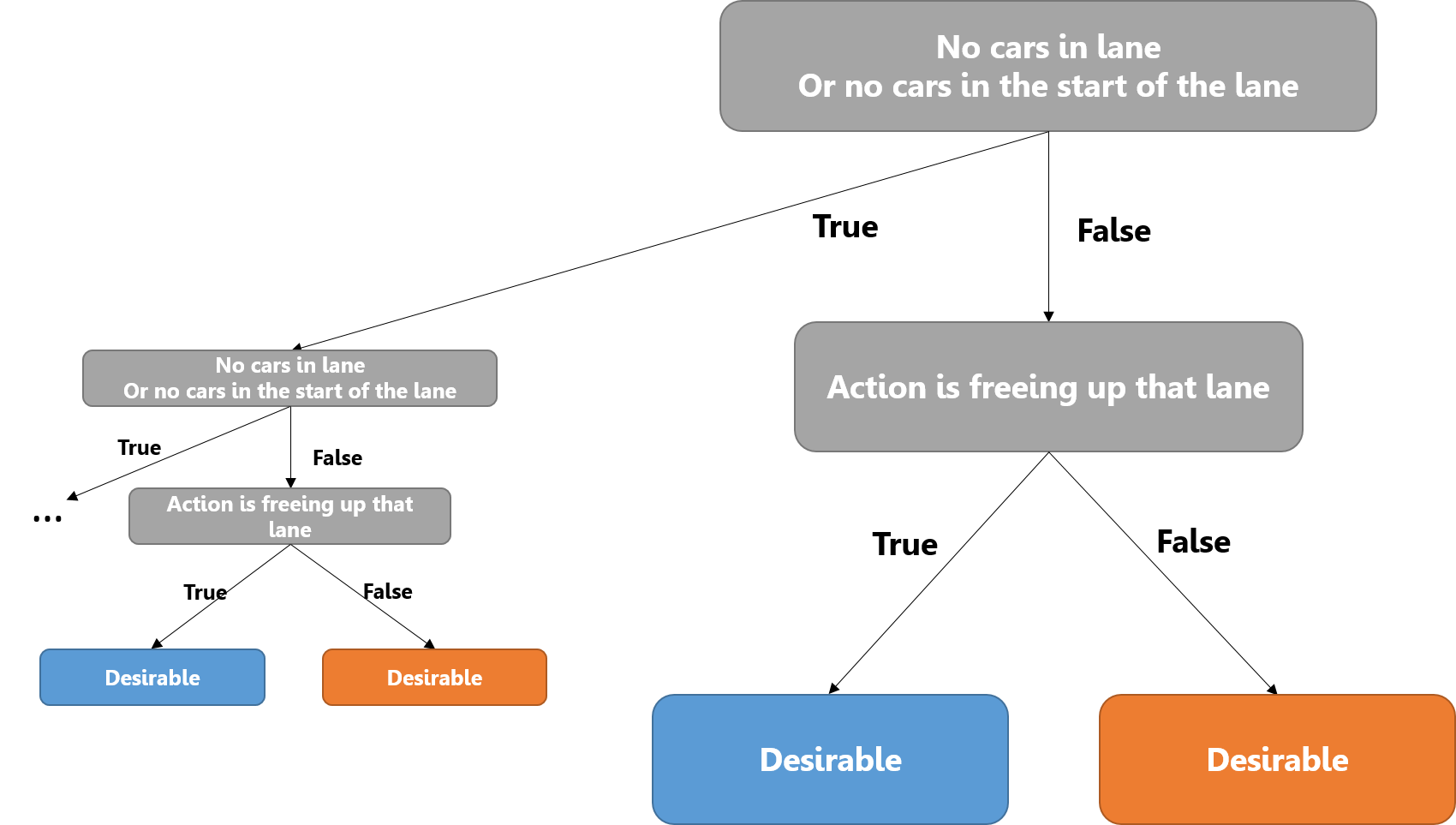}
\caption{Traffic decision tree: diagram of a specific path.}
\label{fig:traffic_path_diagram}
\end{figure}

\medskip
\noindent
\textbf{Traffic Control.}  In Traffic Control, we labeled state-action
pairs as undesirable when the agent failed to take the obvious action
needed to resolve traffic jams.  Our tree achieved over 99\% balanced
accuracy and had depth 10.  Because a tree of depth 10 is difficult to
parse manually, we began by focusing on its top 3 layers, where the
most dominant features reside. In our case, these were the
\emph{Average} and \emph{Action} features. Two interesting paths that
we identified are:
\begin{itemize}
    \item E2W has cars, Action is EW, N2S has fewer than 2 segments filled with cars near the intersection $\rightarrow$ \textbf{Desirable}.
    \item E2W has no cars, N2S has at least one segment filled with cars, Action is not NS $\rightarrow$ \textbf{Undesirable}.
\end{itemize}
These paths in the tree highlight a connection between the first segments of a lane being populated by cars, and the undesirability of not assigning a green light to that lane; and indeed, similar connections appear also in other, deeper parts of the tree. 
For example, when inspecting deeper layers of the tree (up to depth 5), we observed an additional interesting path; see the diagram in Fig.~\ref{fig:traffic_path_diagram}.
The splits along this path examine various lanes in sequence, each time checking whether a lane is populated with cars, at least in its first segments. If this is the case, but the selected action does not assign that lane a green light, this is generally classified as  \emph{undesirable} behavior; and otherwise, the behavior is \emph{desirable}.
This already provides a clear, even if partial, formulation of the property at hand.

We also carried out an additional experiment, in order to assess the effect of different grammars on the interpretability of the resulting tree. Specifically, we took the grammar used for the Snake case study, and used it to train a tree for Traffic Control.
The resulting tree reached 0.98 balanced accuracy, with depth 20, and was generally more difficult to interpret. For example, if we inspect the top 3 layers of the tree, we can indeed infer that the first two segments in a lane are crucial for the outcome of the classification, as the following path demonstrates:
\begin{itemize}
    \item Either N2S or E2W, but not both, have cars in their second segment, ACT is NS, and E2W has cars in its second segment $\rightarrow$ \textbf{Undesirable}.
\end{itemize}
But the conclusion is not as concrete as it was when inspecting the
earlier tree. Ultimately, when inspecting deeper trees, we can come to
similar (but less general) conclusions as with the first tree, but
this process was more difficult to carry out and not as intuitive ---
indicating the importance of picking a grammar that is appropriate for
the system at hand.


\medskip
\noindent
\textbf{Snake.}
In Snake, we labeled state-action pairs as undesirable when the agent made the snake turn away from an apple. We now wish to see whether we can 
 better characterize this  undesirable behavior, using the  tree. For this, we inspect a sub-tree with depth 3, which reaches 94.8\% balanced accuracy (the final tree is of depth 8 and reaches 99.92\% balanced accuracy).

Analyzing the paths of this tree, we observe that the dominant features are ``IsEqual'' features that compare DIR with ACT or with APL. We can deduce from this that DIR is the most important category in determining undesirable behavior. Two of the relevant tree paths are:
\begin{itemize}
\item DIR and ACT are equal in the direction ``DOWN'', DIR and ACT are
  equal in the direction ``RIGHT'', DIR and ACT are equal in the
  direction ``UP'' $\rightarrow$ \textbf{Desirable}.  Because we know
  ACT and DIR are binary vectors of size 4, and that each vector has
  exactly one ``1'' entry and the rest are ``0'', this path implies
  that if DIR == ACT, then the behavior is desirable.  This is indeed
  a key characteristic of the undesirable behavior --- i.e., that DIR
  has to be different from ACT for the behavior to be undesirable.
    
  \item DIR and ACT are not equal in the direction ``DOWN'', DIR and
    APL are equal in the direction ``DOWN'' $\rightarrow$ mostly
    \textbf{Undesirable}.  This path indicates that if DIR and ACT are
    not equal, but DIR and APL are equal, this is mostly undesirable.
\end{itemize}

By looking at this fairly shallow tree, we can make deductions that
assist us in formalizing the undesirable behavior.  By inspecting
deeper trees, we can observe that the classifier is indeed attempting
to figure out the indices in which the DIR, APL, and ACT are equal, in
order to make its classification. We observe that OBS has little
effect here, but that it does appear in the deeper levels of the tree
--- presumably because the cases where an obstacle is nearby are few
and far between.

Next, if we compare this tree to a tree produced using the ``Traffic''
grammar, we see that the tree with the original grammar reaches higher
accuracy, and does so much with more shallow trees. Also, manually
inspecting the tree produced using the ``Traffic'' grammar proved much
more difficult, because the features were less relevant --- for
example, the first split is determined by whether the Average of the
``UP'' and ``RIGHT'' entries in ``APL'' equals 0, which is a
convoluted way of asking whether the direction of APL is either ``UP''
or ``RIGHT''. A deeper inspection revealed that this second tree ended
up making similar decisions to the first, but this took a great deal
of effort.  These results again highlight the importance of picking an
appropriate grammar.

\subsection{Comparison to the State of the Art}
At first glance, it may seem appropriate to compare our method to
reward-reshaping methods for improving the safety of RL-based
systems. We argue, however, that this is not so: by modifying the code
in Listings~\ref{code:reward_modifier} and~\ref{code:modify_reward},
our method can easily be integrated with any reward reshaping method
as a black box, and thus benefit from future improvements in those
techniques. For example, we could adjust our approach to use
potential-based reward reshaping~\cite{NgHaRu99}, by adding an
additive term to the original reward instead of a multiplicative
one. Indeed, our choice of reward reshaping function for our
implementation of the technique is quite arbitrary.



A more relevant comparison is between our approach and other
\emph{Safe-RL methods}, i.e. methods aimed at reducing undesirable
behavior in RL-based systems. Here, we ran into the following
difficulties:
\begin{inparaenum}[(i)]
\item one of the main advantages of our approach is that we do
  not assume an a-priori, rigorous characterization of the undesirable
  behavior we want to reduce, whereas state-of-the-art techniques
  require such a characterization; and
\item the goal of existing Safe-RL techniques is to completely
  eliminate undesirable behavior, whereas our approach allows
  fine-tuning the trade-off between eliminating such behavior and
  maintaining high performance.
\end{inparaenum}
These two issues prevented us from conducting a meaningful comparison
to many of the existing techniques. (As a side note, we point out that
our approach can be integrated with approaches the require a rigorous
characterization of undersealed behavior, by using our trained
decision tree for this purpose.)

The most relevant existing technique, and to which we ended up
comparing our proposed approach, is the VIPER approach proposed by Bastani
et al.~\cite{BaPuSo18}. There, the authors extract a decision tree
policy from a pre-trained DNN. This tree, which is assumed to mimic
the behavior of the DNN fairly well, is then used for verification
purposes.

\begin{figure}[h]
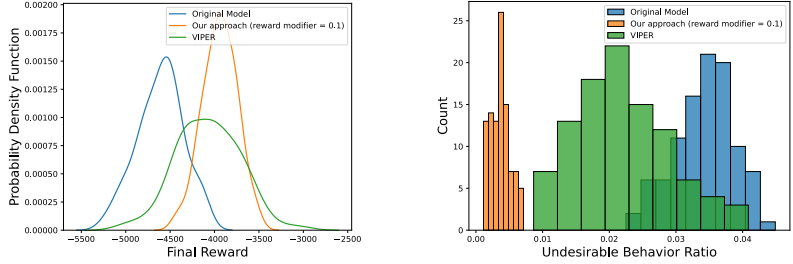

\centering
\includesvg[width=0.4\textwidth]{Images/Traffic/pdf_reward_traffic.svg}
\qquad
\includesvg[width=0.4\textwidth]{Images/Traffic/hist_ratio_traffic.svg}
\caption{Traffic Control: Comparing our approach and VIPER.}
\label{fig:traffic_comparison}
\end{figure}

To compare the approaches, we focused on the Traffic Control use-case,
and performed a comparison between three models:
\begin{inparaenum}[(i)]
\item the original Traffic Control model;
\item the decision tree produced by VIPER; and
\item the model produced by our approach, with the reward modifier
  arbitrarily set to $0.1$, referred to as UBR (for
  undesirable-behavior-reduced).
\end{inparaenum}
%
%
For each of these models, we calculated the undesirable behavior ratio
and the final reward over 100 runs. Figure~\ref{fig:traffic_comparison}
depicts a density plot and a histogram plot comparing the results.

As Fig.~\ref{fig:traffic_comparison} shows, in terms of the reward
values achieved the original model is outperformed by both our model
and the VIPER model; whereas VIPER performs comparably, and even
slightly better, than our approach. In terms of undesirable behavior,
VIPER significantly outperforms the original model, and our model
significantly outperforms VIPER. These results are to be expected, as
our approach reduces the frequency of undesirable behavior, but
achieves this by altering the training process, possibly reducing the
achieved reward.

Apart from these criteria, we also set out to evaluate the
explainability afforded by our approach, compared to VIPER.  The VIPER
tree had 3139 nodes, whereas the UBR tree had only 511 nodes. Further,
the VIPER tree is trained on the state space (in this case, 80 binary
values), as opposed to our tree that is trained using a user-defined
grammar, and consequently its tree nodes were not as uninformative.
Inspecting the different paths of the VIPER tree for depth 3, we
observed the following:
\begin{itemize}
    \item W2E[0] = 0 $\rightarrow$ E2W[0] = 0 $\rightarrow$ E2TL[0] = 0 $\rightarrow$ EWL
    
    \item W2E[0] = 0 $\rightarrow$ E2W[0] = 0 $\rightarrow$ E2TL[0] = 1 $\rightarrow$ NSL
    
    \item W2E[0] = 0 $\rightarrow$ E2W[0] = 1 $\rightarrow$ E2TL[0] = 0 $\rightarrow$ NS
    
    \item W2E[0] = 0 $\rightarrow$ E2W[0] = 1 $\rightarrow$ E2TL[0] = 1 $\rightarrow$ NSL
    
    \item W2E[0] = 1 $\rightarrow$ E2W[0] = 0 $\rightarrow$ S2TL[0] = 0 $\rightarrow$ EW
    
    \item W2E[0] = 1 $\rightarrow$ E2W[0] = 0 $\rightarrow$ S2TL[0] = 1 $\rightarrow$ NSL
    
    \item W2E[0] = 1 $\rightarrow$ E2W[0] = 1 $\rightarrow$ S2N[1] = 0 $\rightarrow$ EW
    
    \item W2E[0] = 1 $\rightarrow$ E2W[0] = 1 $\rightarrow$ S2N[1] = 1 $\rightarrow$ NS
\end{itemize}
Clearly, these paths indicate that the first indices of each lane are
the most dominant, and that the selected action roughly corresponds to
the lane in which there are cars. However, by just inspecting these
paths we cannot fully explain the selected actions --- especially when
considering that the state-space is rich, and that selecting an action
based strictly on which lanes have cars is suboptimal. Delving deeper
into the tree affords a better understanding, but this required
significantly more work than with the UBR tree.

\begin{figure}[h]
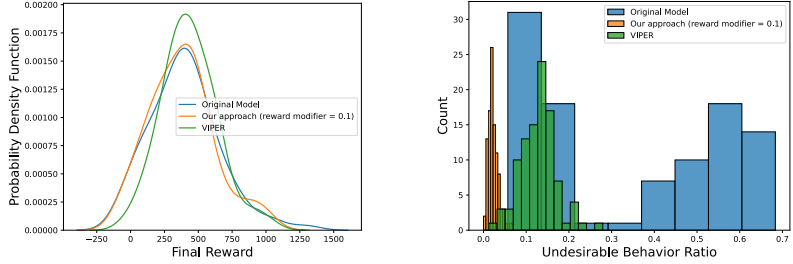

\centering
\includesvg[width=0.4\textwidth]{Images/snake/pdf_reward_snake.svg}
\qquad
\includesvg[width=0.4\textwidth]{Images/snake/hist_ratio_snake.svg}
\caption{Snake: Comparing our approach and VIPER.}
\label{fig:snake_comparison}
\end{figure}

Next, we repeated the experiment with the Snake case-study, comparing
the original model (using an average of 4 models), our
undesirable-behavior-reduced (UBR) model (trained with reward modifier
0.1, using an average of 4 models), and the VIPER model.
We calculated the undesirable behavior ratio and the final reward for
100 runs for all models. The results appear in
Fig.~\ref{fig:snake_comparison}.

As we can see in Fig.~\ref{fig:snake_comparison}, all 3 models perform
similarly to each other, with VIPER very slightly outperforming the
original model, and our UBR model slightly outperforming both. Also,
VIPER has a slightly lower undesirability ratio than the original
model, and the UBR model has the lowest undesirability ratio.

Comparing the explainability of the VIPER and UBR trees, we again
observe that UBR affords better explainability. Specifically, the UBR
tree has fewer nodes --- 211, compared to 241 in the VIPER
tree. Inspecting the paths of the VIPER tree up to depth 3, we get:
\begin{itemize}
    \item DIRECTION is not RIGHT $\rightarrow$ DIRECTION is not LEFT $\rightarrow$ APPLE is not DOWN $\rightarrow$ LEFT

    \item DIRECTION is not RIGHT $\rightarrow$ DIRECTION is not LEFT $\rightarrow$ APPLE is DOWN $\rightarrow$ LEFT
    
    \item DIRECTION is not RIGHT $\rightarrow$ DIRECTION is LEFT $\rightarrow$ APPLE is not DOWN $\rightarrow$ DOWN
    
    \item DIRECTION is not RIGHT $\rightarrow$ DIRECTION is LEFT $\rightarrow$ APPLE is DOWN $\rightarrow$ LEFT
    
    \item DIRECTION is RIGHT $\rightarrow$ APPLE is not RIGHT $\rightarrow$ APPLE is not UP $\rightarrow$ LEFT
    
    \item DIRECTION is RIGHT $\rightarrow$ APPLE is not RIGHT $\rightarrow$ APPLE is UP $\rightarrow$ UP
    
    \item DIRECTION is RIGHT $\rightarrow$ APPLE is RIGHT $\rightarrow$ OBSTACLE is not RIGHT $\rightarrow$ RIGHT

    \item DIRECTION is RIGHT $\rightarrow$ APPLE is RIGHT $\rightarrow$ OBSTACLE is RIGHT $\rightarrow$ LEFT
    
\end{itemize}
We observe that the most dominant features are DIRECTION and APPLE,
similarly to the UBR tree. However, other than the paths ``DIRECTION
is RIGHT $\rightarrow$ APPLE is RIGHT $\rightarrow$ OBSTACLE is not
RIGHT $\rightarrow$ RIGHT''", and ``DIRECTION is RIGHT $\rightarrow$
APPLE is not RIGHT $\rightarrow$ APPLE is UP $\rightarrow$ UP'', in
which the action is to move towards the apple, it is not intuitively
clear how decisions are made.  It also seems that the ``default''
choice is to turn LEFT, even though this can be suboptimal. Upon a
deeper inspection longer paths, we can infer additional behaviors of
the tree, but this is not intuitive.

We conclude that in both case-studies, we were able to train a VIPER
model that adequately mimics the original model --- but also that in
both cases, the UBR approach achieved higher rewards than VIPER, and
afforded superior explainability. This is perhaps not surprising, as
VIPER is geared towards verification, and not towards
interpretability.


\section{Related Work} \label{sec:related}
\subsection{Related Work}
Reducing undesirable behavior in deep reinforcement learning models
has been studied extensively, mostly within the field
of \textit{Safe Reinforcement
  Learning}~\cite{GuYaDuChWaWaYaKn23}. Tessler et al.~\cite{TeMaMa18}
propose an approach that uses an actor-critic method to penalize the
agent's reward, in order to change the policy being learned into one
that satisfies various constraints. Zhang and Guo~\cite{ZhGu22}
suggest a risk-preventative training method, which utilizes a
classifier that predicts the risk of traces becoming unsafe, and
penalizes the reward accordingly --- in order to prevent risky
behavior. Dalal et al.~\cite{Da18} propose to add a safety layer that
corrects selected actions into the closest action that does not lead
to a safety violation, and in that way prevent the agent from ever
reaching unsafe states during training and after deployment.

Our approach is similar to the aforementioned ones, in that it
attempts to reduce undesirable behaviors prevalent in the
model. However, our approach is less restrictive. For example, we do
not attempt to eliminate the undesirable behavior entirely, instead
allowing the user to decide on the trade-off between reducing
undesirable behavior and reducing performance (as we saw, e.g., in
Case Study~\ref{case:Snake}). Also, existing approaches often assume
that the undesirable behavior is well specified, as a hard safety
constraint. In our approach, we circumvent this requirement, and only
assume that a human engineer flags undesirable behavior, without
necessarily understanding the underlying causes.

Thomas et al.~\cite{ThCaBaGiBrBr19} tackle a similar goal but from a
different angle, and design a machine-learning framework that uses a
``Seldonian optimization'' approach in order to prevent undesirable
behavior specified by the user.  Unlike our framework, this approach
does not afford improved explainability, but it does demonstrate the
usefulness of a user-provided flagging of undesirable behavior ---
which we consider as encouraging evidence of the potential of this
line of research.

Bastani et al.~\cite{BaPuSo18} train a decision tree based on the
DRL's labeling, and try to replicate the DRL policy with the
decision tree. This allows much easier verification, as a tree is much
simpler to verify than a DNN.  We explored the similarities and
differences to our approach in the previous section.

\section{Conclusion and Future Work} \label{sec:conc}
The increasing pervasiveness of DRL poses new challenges when it
comes to safety and explainability. We presented here an approach
aimed at rendering DRL systems safer, by identifying undesired
behaviors in their traces and then using this information, via reward
reshaping, to improve their training. In addition to increased safety,
our techniques also serve to increase the explainability of these
systems. Our evaluation on three diverse and significant case studies
indicates the great potential of this line of work.  The main novelty
of our approach is that, in contrast to Safe-RL methods, it does 
not assume a rigorous, a-priori characterization of undesirable
behavior. Instead, we learn the undesirable behavior from user input,
and then retrain the model in order to reduce its undesirable behavior.

Moving forward, there are several directions we plan to explore.  One
such direction is to try and generalize our method to entire traces of
undesirable/desirable behavior, as opposed to individual state-action
pairs. This would allow our learned decision trees to affect the
reward function in more subtle ways, and hopefully result in safer
systems.  Another direction we plan to pursue is to create a
double-blind trial of classifying undesirable behavior and
interpreting the tree, in order to provide further credence to our
approach.  Finally, we plan to explore the option of replacing the
decision trees used as part of our approach with other
software-engineering constructs, which may be more amenable for manual
inspection and analysis when the system at hand becomes complex. One
such software-engineering construct is that of scenario-based
programming~\cite{HaKa14,HaMaWe12ACM}, which has been shown to produce
software artifacts that are in line with how humans perceive complex
systems~\cite{AlArGoHa14,GoMaMe12Spaghetti,HaKaLaMaWe15,HaKaMaMa16,HaKaMaMa18},
and which may be automatically analyzed, adjusted and
extended~\cite{HaKaKa13,HaKaMaWe14,Ka13}.

{
\bibliographystyle{abbrv}
\bibliography{bib}
}

\end{document}